\documentclass[10pt,twocolumn,letterpaper]{article}

\usepackage{iccv}
\usepackage{times}
\usepackage{epsfig}
\usepackage{graphicx}
\usepackage{amsmath}
\usepackage{amssymb}
\usepackage{multirow}
\usepackage{bm}
\usepackage{amsmath}
\usepackage[pagebackref=true,breaklinks=true,letterpaper=true,colorlinks,bookmarks=false]{hyperref}

 \iccvfinalcopy % *** Uncomment this line for the final submission

\def\iccvPaperID{638} % *** Enter the ICCV Paper ID here

\ificcvfinal\fi
\begin{document}

%%%%%%%%% TITLE
\title{VQABQ: Visual Question Answering by Basic Questions}

\author{Jia-Hong Huang~~~~~~~~~~Modar Alfadly~~~~~~~~~~Bernard Ghanem\\
King Abdullah University of Science and Technology\\
%KAUST EE, Thuwal, Saudi Arabia\\
{\tt\small \{jiahong.huang, modar.alfadly, bernard.ghanem\}@kaust.edu.sa}
%\and
%Modar Alfadly\\
%King of Abdullah University of Science and Technology\\
%{\tt\small modar.alfadly@kaust.edu.sa}
%\and
%Bernard Ghanem\\
%King of Abdullah University of Science and Technology\\
%First line of institution3 address\\
%{\tt\small bernard.ghanem@kaust.edu.sa}
}

\maketitle
%\thispagestyle{empty}

%%%%%%%%% ABSTRACT
\begin{abstract}
   Taking an image and question as the input of our method, it can output the text-based answer of the query question about the given image, so called Visual Question Answering (VQA). There are two main modules in our algorithm. Given a natural language question about an image, the first module takes the question as input and then outputs the basic questions of the main given question. The second module takes the main question, image and these basic questions as input and then outputs the text-based answer of the main question. We formulate the basic questions generation problem as a LASSO optimization problem, and also propose a criterion about how to exploit these basic questions to help answer main question. Our method is evaluated on the challenging VQA dataset \cite{4} and yields state-of-the-art accuracy, 60.34\% in open-ended task. 
\end{abstract}

%%%%%%%%% BODY TEXT
\section{Introduction}

Visual Question Answering (VQA) is a challenging and young research field, which can help machines achieve one of the ultimate goals in computer vision, holistic scene understanding \cite{50}. VQA is a computer vision task: a system is given an arbitrary text-based question about an image, and then it should output the text-based answer of the given question about the image. The given question may contain many sub-problems in computer vision, e.g., 

\begin{figure}[t]
\begin{center}
 \includegraphics[width=1.005\linewidth]{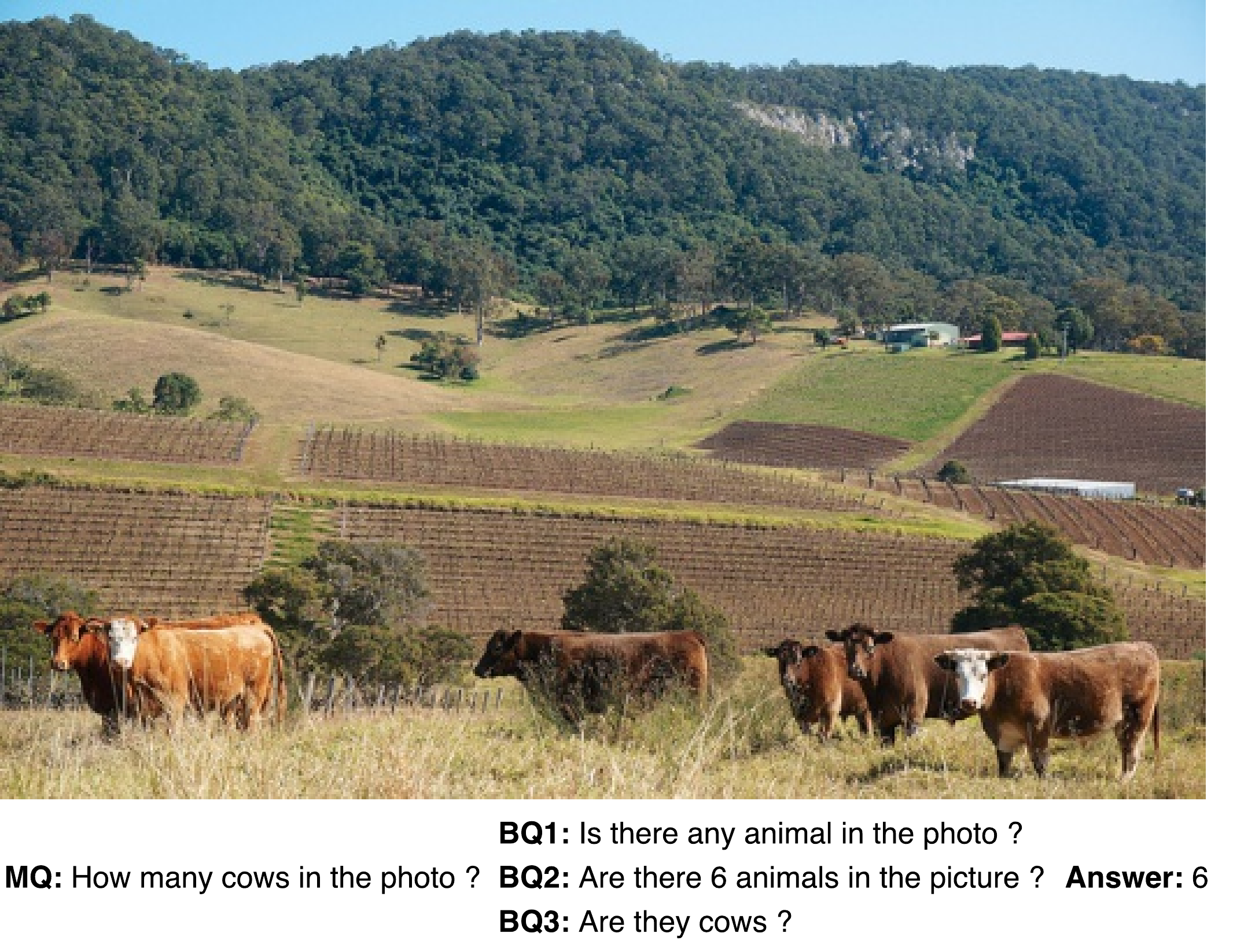}
\end{center}
   \caption{Examples of basic questions. Note that MQ denotes the main question and BQ denotes the basic question.}
\label{fig:figure1}
\end{figure}

\begin{itemize}
   \item Scene classification - Is it a rainy day?
\end{itemize}
    
\begin{itemize}
   \item Object recognition - What is on the desk? 
\end{itemize}

\begin{itemize}
   \item Attribute classification - What color is the ground?
\end{itemize}

\begin{itemize}
   \item Counting - How many people are in the room?
\end{itemize}

\begin{itemize}
   \item Object detection - Are there any apples in the image?
\end{itemize}
 
\begin{itemize}
   \item Activity recognition - What kind of exercise is the man doing?
\end{itemize}

Besides, in our real life there are a lot of more complicated questions that can be queried. So, in some sense, VQA can be considered as an important basic research problem in computer vision. From the above sub-problems in computer vision, we can discover that if we want to do holistic scene understanding in one step, it is probably too difficult. So, we try to divide the holistic scene understanding-task into many sub-tasks in computer vision. The task-dividing concept inspires us to do Visual Question Answering by Basic Questions (VQABQ), illustrated by Figure \ref{fig:figure1}. That means, in VQA, we can divide the query question into some basic questions, and then exploit these basic questions to help us answer the main query question. 
Since 2014, there has been a lot of progress in designing systems with the VQA ability \cite{2, 4, 9, 12, 14, 33}.  Regarding these works, we can consider most of them as visual-attention VQA works because most of them do much effort on dealing with the image part but not the text part.  
However, recently there are some works \cite{41, 42} that try to do more effort on the question part. In \cite{42}, authors proposed a Question Representation Update (QRU) mechanism to update the original query question to increase the accuracy of the VQA algorithm. Typically, VQA is a strongly image-question dependent issue, so we should pay equal attention to both the image and question, not only one of them. In reality, when people have an image and a given question about the image, we usually notice the keywords of the question and then try to focus on some parts of the image related to question to give the answer. So, paying equal attention to both parts is a more reasonable way to do VQA. In \cite{41}, the authors proposed a Co-Attention mechanism, jointly utilizing information about visual and question attention, for VQA and achieved the state-of-the-art accuracy.

\begin{figure}
\begin{center}
   \includegraphics[width=0.9\linewidth]{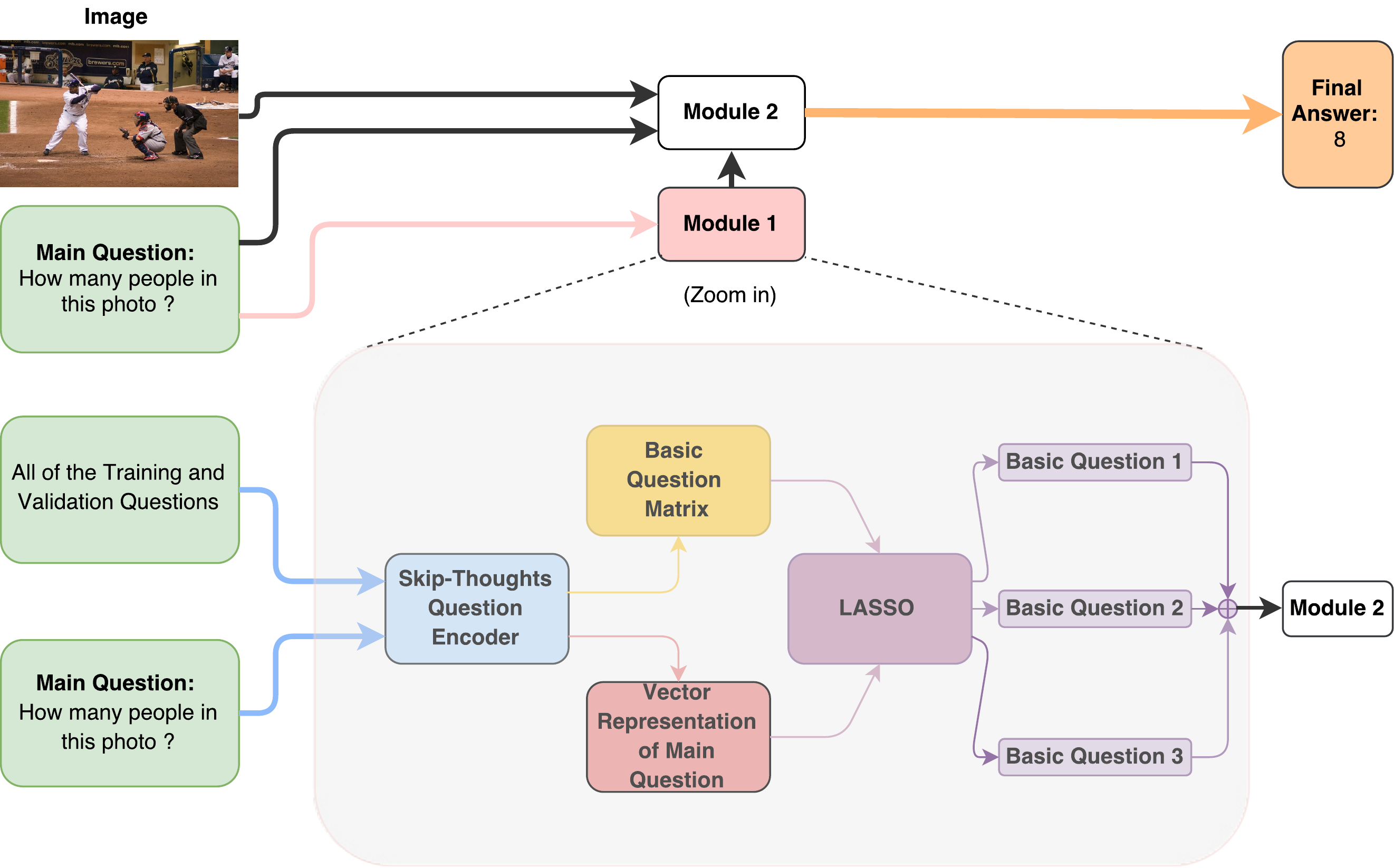}
\end{center}
   \caption{VQABQ working pipeline. Note that all of the training and validation questions are only encoded by Skip-Thoughts one time for generating the basic question matrix. That is, the next input of Skip-Thoughts is only the new main question. Here, "$\oplus$" denotes the proposed basic question concatenation method.}
\label{fig:figure2}
\end{figure}

The Co-Attention mechanism inspires us to build part of our VQABQ model, illustrated by Figure \ref{fig:figure2}. In the VQABQ model, there are two main modules, the basic question generation module (Module 1) and co-attention visual question answering module (Module 2). We take the query question, called the main question (MQ), encoded by Skip-Thought Vectors \cite{43}, as the input of Module 1. In the Module 1, we encode all of the questions, also by Skip-Thought Vectors, from the training and validation sets of VQA \cite{4} dataset as a 4800 by 215623 dimension basic question (BQ) matrix, and then solve the LASSO optimization problem, with MQ, to find the 3 BQ of MQ. These BQ are the output of Module 1. Moreover, we take the MQ, BQ and the given image as the input of Module 2, the VQA module with co-attention mechanism, and then it can output the final answer of MQ. We claim that the BQ can help Module 2 get the correct answer to increase the VQA accuracy. In this work, our main contributions are summarized below: 

\bigskip
\begin{itemize}
    \item We propose a method to generate the basic questions of the main question and utilize these basic questions with proper criterion to help answer the main question in VQA.
    \item Also, we propose a new basic question dataset generated by our basic question generation algorithm. 
\end{itemize}

\noindent
The rest of this paper is organized as the following. We first talk about the motivation about this work in Section 2. In Section 3, we review the related work, and then Section 4 shortly introduces the proposed VQABQ dataset. We discuss the detailed methodology in Section 5. Finally, the experimental results are demonstrated in Section 6.
%-------------------------------------------------------------------------
\section{Motivations}

The following two important reasons motivate us to do Visual Question Answering by Basic Questions (VQABQ). First, recently most of VQA works only emphasize more on the image part, the visual features, but put less effort on the question part, the text features. However, image and question features both are important for VQA. If we only focus on one of them, we probably cannot get the good performance of VQA in the near future. Therefore, we should put our effort more on both of them at the same time. In \cite{41}, they proposed a novel co-attention mechanism that jointly performs image-guided question attention and question-guided image attention for VQA. \cite{41} also proposed a hierarchical architecture to represent the question, and construct image-question co-attention maps at the word level, phrase level and question level. Then, these co-attended features are combined with word level, phrase level and question level recursively for predicting the final answer of the query question based on the input image. \cite{42} is also a recent work focusing on the text-based question part, text feature. In \cite{42}, they presented a reasoning network to update the question representation iteratively after the question interacts with image content each time. Both of \cite{41, 42} yield better performance than previous works by doing more effort on the question part. 

Secondly, in our life , when people try to solve a difficult problem, they usually try to divide this problem into some small basic problems which are usually easier than the original problem. So, why don't we apply this dividing concept to the input question of VQA ? If we can divide the input main question into some basic questions, then it will help the current VQA algorithm achieve higher probability to get the correct answer of the main question.

Thus, our goal in this paper is trying to generate the basic questions of the input question and then exploit these questions with the given image to help the VQA algorithm get the correct answer of the input question. Note that we can consider the generated basic questions as the extra useful information to VQA algorithm.

\section{Related Work}

Recently, there are many papers \cite{4, 13, 15, 18, 30, 32, 34, 37} have proposed methods to solve the VQA issue. Our method involves in different areas in machine learning, natural language processing (NLP) and computer vision. The following, we discuss recent works related to our approach for solving VQA problem. 
\bigskip

\noindent
\textbf{Sequence modeling by Recurrent Neural Networks.} 

Recurrent Neural Networks (RNN) can handle the sequences of flexible length. Long Short Term Memory (LSTM) \cite{44} is a particular variant of RNN and in natural language tasks, such as machine translation \cite{45, 46}, LSTM is a successful application. In \cite{32}, the authors exploit RNN and Convolutional Neural Network (CNN) to build a question generation algorithm, but the generated question sometimes has invalid grammar. The input in \cite{9} is the concatenation of each word embedding with the same feature vector of image. \cite{33} encodes the input question sentence by LSTM and join the image feature to the final output. \cite{30} groups the neighbouring word and image features by doing convolution. In \cite{6}, the question is encoded by Gated Recurrent Unit (GRU) \cite{10} similar to LSTM and the authors also introduce a dynamic parameter layer in CNN whose weights are adaptively predicted by the encoded question feature.   
\bigskip

\noindent
\textbf{Sentence encoding.}

In order to analyze the relationship among words, phrases and sentences, several works, such as \cite{11, 43, 47}, proposed methods about how to map text into vector space. After we have the vector representation of text, we can exploit the vector analysis skill to analyze the relationship among text. \cite{11, 47} try to map words to vector space, and if the words share common contexts in the corpus, their encoded vectors will close to each other in the vector space. In \cite{43}, the authors propose a framework of encoder-decoder models, called skip-thoughts. In this model, the authors exploit an RNN encoder with GRU activations \cite{10} and an RNN decoder with a conditional GRU \cite{10}. Because skip-thoughts model emphasizes more on whole sentence encoding, in our work, we encode the whole question sentences into vector space by skip-thoughts model and use these skip-thought vectors to do further analysis of question sentences.

\bigskip

\noindent
\textbf{Image captioning.}

In some sense, VQA is related to image captioning \cite{1, 27, 28, 48}. \cite{48} uses a language model to combine a set of possible words detected in several regions of the image and generate image description. In \cite{28}, the authors use CNN to extract the high-level image features and considered them as the first input of the recurrent network to generate the caption of image. \cite{1} proposes an algorithm to generate one word at a time by paying attention to local image regions related to the currently predicted word. In \cite{27}, the deep neural network can learn to embed language and visual information into a common multi-modal space. However, the current image captioning algorithms only can generate the rough description of image and there is no so called proper metric to evaluate the quality of image caption , even though BLEU \cite{49} can be used to evaluate the image caption.

\bigskip

\noindent
\textbf{Attention-based VQA.}

There are several VQA models have ability to focus on specific image regions related to the input question by integrating the image attention mechanism \cite{13, 15, 35, 42}. In \cite{42}, in the pooling step, the authors exploit an image attention mechanism to help determine the relevance between original questions and updated ones. Before \cite{41}, no work applied language attention mechanism to VQA, but the researchers in NLP they had modeled language attention. In \cite{41}, the authors propose a co-attention mechanism that jointly performs language attention and image attention. Because both question and image information are important in VQA, in our work we introduce co-attention mechanism into our VQABQ model.

\section{Basic Question Dataset}
We propose a new dataset, called Basic Question Dataset (BQD), generated by our basic question generation algorithm. BQD is the first basic question dataset. Regarding the BQD, the dataset format is $\{Image,~MQ,~3~(BQ + corresponding~similarity~score)\}$. All of our images are from the testing images of MS COCO dataset \cite{51}, the MQ, main questions, are from the testing questions of VQA, open-ended, dataset \cite{4}, the BQ, basic questions, are from the training and validation questions of VQA, open-ended, dataset \cite{4}, and the corresponding similarity score of BQ is generated by our basic question generation method, referring to Section 5. Moreover, we also take the multiple-choice questions in VQA dataset \cite{4} to do the same thing as above. Note that we remove the repeated questions in the VQA dataset, so the total number of questions is slightly less than VQA dataset \cite{4}. In BQD, we have 81434 images, 244302 MQ and 732906 (BQ + corresponding similarity score). At the same time, we also exploit BQD to do VQA and achieve the competitive accuracy compared to state-of-the-art.
%-------------------------------------------------------------------------
\section{Methodology}
In Section 5, we mainly discuss how to encode questions and generate BQ and why we exploit the Co-Attention Mechanism VQA algorithm \cite{41} to answer the query question. The overall architecture of our VQABQ model can be referred to Figure \ref{fig:figure2}. The model has two main parts, Module 1 and Module 2. Regarding Module 1, it takes the encoded MQ as input and uses the matrix of the encoded BQ to output the BQ of query question. Then, the Module 2 is a VQA algorithm with the Co-Attention Mechanism \cite{41}, and it takes the output of Module 1, MQ, and the given image as input and then outputs the final answer of MQ. The detailed architecture of Module 1 can be referred to Figure \ref{fig:figure2}.

\subsection{Question encoding}

\noindent

There are many popular text encoders, such as Word2Vec \cite{47}, GloVe \cite{11} and Skip-Thoughts \cite{43}. In these encoders, Skip-Thoughts not only can focus on the word-to-word meaning but also the whole sentence semantic meaning. So, we choose Skip-Thoughts to be our question encoding method. In Skip-Thoughts model, it uses an RNN encoder with GRU \cite{10} activations, and then we use this encoder to map an English sentence into a vector. Regarding GRU, it has been shown to perform as well as LSTM \cite{44} on the sequence modeling applications but being conceptually simpler because GRU units only have 2 gates and do not need the use of a cell.

\bigskip

\noindent
\textbf{Question encoder.} Let $w_{i}^{1},...,w_{i}^{N}$ be the words in question $s_{i}$ and N is the total number of words in $s_{i}$. Note that $w_{i}^{t}$ denotes the $t$-th word for $s_{i}$ and $\mathbf{x}_{i}^t$ denotes its word embedding. The question encoder at each time step generates a hidden state $\mathbf{h}_{i}^{t}$. It can be considered as the representation of the sequence $w_{i}^{1},..., w_{i}^{t}$. So, the hidden state $\mathbf{h}_{i}^{N}$ can represent the whole question. For convenience, here we drop the index $i$ and iterate the following sequential equations to encode a question:
\begin{equation}
    \mathbf{r}^{t}~=~\sigma (\mathbf{U}_{r}\mathbf{h}^{t-1}+\mathbf{W}_{r}\mathbf{x}^{t})
\end{equation}
\begin{equation}
    \mathbf{z}^{t}~=~\sigma(\mathbf{U}_{z}\mathbf{h}^{t-1}+\mathbf{W}_{z}\mathbf{x}^{t})
\end{equation} 
\begin{equation}
    \bar{\mathbf{h}}^{t}~=~\mathrm{tanh}(\mathbf{U}(\mathbf{r}^{t}\odot \mathbf{h}^{t-1})+\mathbf{Wx}^{t})
\end{equation} 
\begin{equation}
    \mathbf{h}^{t}~=~\mathbf{z}^{t}\odot  \bar{\mathbf{h}}^{t}+(1-\mathbf{z}^{t})\odot \mathbf{h}^{t-1}
\end{equation}

\noindent
, where $\mathbf{U}_{r}$, $\mathbf{U}_{z}$, $\mathbf{W}_{r}$, $\mathbf{W}_{z}$, $\mathbf{U}$ and $\mathbf{W}$ are the matrices of weight parameters. $\bar{\mathbf{h}}^{t}$ is the state update at time step $t$, $\mathbf{r}^{t}$ is the reset gate, $\odot$ denotes an element-wise product and $\mathbf{z}^{t}$ is the update gate. These two update gates take the values between zero and one.

% \begin{figure*}
% \begin{center}
% \includegraphics[width=0.9\linewidth]{Module1_pipeline.png}
% \end{center}
%   \caption{Basic Question Generation Module, (Module 1), working pipeline.}
% \label{fig:short}
% \end{figure*}

\subsection{Problem Formulation}

\noindent

Our idea is the BQ generation for MQ and, at the same time, we only want the minimum number of BQ to represent the MQ, so modeling our problem as $LASSO$ optimization problem is an appropriate way:

\begin{equation}
    \min_{\mathbf{x}}~\frac{1}{2}\left \| A\mathbf{x}-\mathbf{b} \right \|_{2}^{2}+\lambda \left \| \mathbf{x} \right \|_{1}
\end{equation}
, where $A$ is the matrix of encoded BQ, $\mathbf{b}$ is the encode MQ and $\lambda$ is a parameter of the regularization term. 

\subsection{Basic Question Generation}

\noindent

We now describe how to generate the BQ of a query question, illustrated by Figure \ref{fig:figure2}. Note that the following we only describe the open-ended question case because the multiple-choice case is same as open-ended one. According to Section 5.2, we can encode the all questions from the training and validation questions of VQA dataset \cite{4} by Skip-Thought Vectors, and then we have the matrix of these encoded basic questions. Each column of the matrix is the vector representation, 4800 by 1 dimensions, of a basic question and we have 215623 columns. That is, the dimension of BQ matrix, called $A$, is 4800 by 215623. Also, we encode the query question as a column vector, 4800 by 1 dimensions, by Skip-Thought Vectors, called $\mathbf{b}$. Now, we can solve the $LASSO$ optimization problem, mentioned in Section 5.3, to get the solution, $\mathbf{x}$. Here, we consider the elements, in solution vector $\mathbf{x}$, as the weights of the corresponding BQ in BQ matrix, $A$. The first element of $\mathbf{x}$ corresponds to the first column, i.e. the first BQ, of $A$. Then, we rank the all weights in $\mathbf{x}$ and pick up the top 3 large weights with corresponding BQ to be the BQ of the query question. Intuitively, because BQ are important to MQ, the weights of BQ also can be considered as importance scores and the BQ with larger weight means more important to MQ. Finally, we find the BQ of all 142093 testing questions from VQA dataset and collect them together, with the format $\{Image,~MQ,~3~(BQ + corresponding~ similarity~score)\}$, as the BQD in Section 4.

\subsection{Basic Question Concatenation}

\noindent

In this section, we propose a criterion to use these BQ. In BQD, each MQ has three corresponding BQ with scores. We can have the following format, $\{MQ,(BQ1,~score1),(BQ2,~score2),(BQ3,~score3)\}$, and these scores are all between 0 and 1 with the following order, 
\begin{equation}
    score1\geq score2\geq score3
\end{equation}
and we define 3 thresholds, $s1$, $s2$ and $s3$. Also, we compute the following 3 averages ($avg$) and 3 standard deviations ($std$) to $score1$, $score2/score1$ and $score3/score2$, respectively, and then use $avg \pm std$, referring to Table \ref{table:table3}, to be the initial guess of proper thresholds. The BQ utilization process can be explained as Table \ref{table:table1}. The detailed discussion about BQ concatenation algorithm is described in the Section 6.4.
\begin{center}
\begin{table}
\centering
\begin{tabular}{llll}
\hline
\multicolumn{4}{c}{\textbf{Basic Question Concatenation Algorithm}}\\ \hline  
\multicolumn{4}{l}
{\multirow{7}{*}
{\begin{tabular}[c]{@{}l@{}}Note that s1, s2, s3 are thresholds we can choose.\\ 
1:~~~~~~\textbf{if} score1 \textgreater s1\\ 
2:~~~~~~~~~Append BQ1 with the largest score\\ 
3:~~~~~~~~~\textbf{if} score2/score1 \textgreater s2\\ 
4:~~~~~~~~~~~~Append BQ2 with the second large score\\
5:~~~~~~~~~~~~\textbf{if} score3/score2 \textgreater s3\\ 
6:~~~~~~~~~~~~~~~Append BQ3 with the third large score\\ 
7:~~~~~~~~~~~~\textbf{else}\\ 
8:~~~~~~~~~~~~~~~None\\ 
9:~~~~~~~~~\textbf{else} \\ 
10:~~~~~~~~~~~None\\ 
11:~~~~~\textbf{else}\\ 
12:~~~~~~~~None
\end{tabular}}}\\
\multicolumn{4}{l}{}\\
\multicolumn{4}{l}{}\\
\multicolumn{4}{l}{}\\
\multicolumn{4}{l}{}\\
\multicolumn{4}{l}{}\\
\multicolumn{4}{l}{}\\
\multicolumn{4}{l}{}\\
\multicolumn{4}{l}{}\\
\multicolumn{4}{l}{}\\
\multicolumn{4}{l}{}\\
\multicolumn{4}{l}{}\\
\multicolumn{4}{l}{}\\
\hline
\end{tabular}
\caption{Note that appending BQ means doing the concatenation with MQ.}
\label{table:table1}
\end{table}
\end{center}

\subsection{Co-Attention Mechanism}

\noindent

There are two types of Co-Attention Mechanism \cite{41} , Parallel and Alternating. In our VQABQ model, we only use the VQA algorithm with Alternating Co-Attention Mechanism to be our VQA module, referring to Figure \ref{fig:figure2}, because, in \cite{41}, Alternating Co-Attention Mechanism VQA module can get the higher accuracy than the Parallel one. Moreover, we want to compare with the VQA method, Alternating one, with higher accuracy in \cite{41}. In Alternating Co-Attention Mechanism, it sequentially alternates between generating question and image attention. That is, this mechanism consists of three main steps: 
\begin{itemize}
   \item First, the input question is summarized into a single vector $\bm{q}$.
   \item Second, attend to the given image depended on $\bm{q}$.  
   \item Third, attend to the question depended on the attended image feature.
\end{itemize}

We can define $\hat{\mathbf{x}}$ is an attention operator, which is a function of $\mathbf{X}$ and $\bm{g}$. This operator takes the question (or image) feature $\mathbf{X}$ and attention guider $\bm{g}$ derived from image (or question) as inputs, and then outputs the attended question (or image) vector. We can explain the above operation as the following steps:

\begin{equation}
    \mathbf{H}~=~\rm{tanh}(\mathbf{W}_{x}\mathbf{X}+(\mathbf{W}_{\pmb{g}}\pmb{g})\mathbf{1}^{T})
\end{equation}
\begin{equation}
    \mathbf{a}^{x}~=~\rm{softmax}(\mathbf{w}_{hx}^{T}\mathbf{H})
\end{equation}
\begin{equation}
    \hat{\mathbf{x}}~=~\sum a_{i}^{x}\mathbf{x}_{i}
\end{equation}
, where $\mathbf{a}^{x}$ is the attention weight of feature $\mathbf{X}$, $\mathbf{1}$ is a vector whose elements are all equal to 1, and $\mathbf{W}_{\pmb{g}}$, $\mathbf{W}_{x}$ and $\mathbf{w}_{hx}$ are matrices of parameters.

Concretely, at the first step of Alternating Co-Attention Mechanism, $\bm{g}$ is $0$ and $\mathbf{X} = \mathbf{Q}$. Then, at the second step, $\mathbf{X} = \mathbf{V}$ where $\mathbf{V}$ is the image features and the guider, $\bm{g}$, is intermediate attended question feature, $\hat{s}$, which is from the first step. At the final step, it uses the attended image feature, $\hat{v}$, as the guider to attend the question again. That is, $\mathbf{X} = \mathbf{Q}$ and $\bm{g} = \hat{v}$.

\section{Experiment}

In Section 6, we describe the details of our implementation and discuss the experiment results about the proposed method. 

\subsection{Datasets}

We conduct our experiments on VQA \cite{4} dataset. VQA dataset is based on the MS COCO dataset \cite{51} and it contains the largest number of questions. There are questions, 248349 for training, 121512 for validation and 244302 for testing. In the VQA dataset, each question is associated with 10 answers annotated by different people from Amazon Mechanical Turk (AMT). About 98\% of answers do not exceed 3 words and 90\% of answers have single words. Note that we only test our method on the open-ended case in VQA dataset because it has the most open-ended questions among the all available dataset and we also think open-ended task is closer to the real situation than multiple-choice one.

\subsection{Setup}

\noindent

In order to prove our claim that BQ can help accuracy and compare with the state-of-the-art VQA method \cite{41}, so, in our Module 2, we use the same setting, dataset and source code mentioned in \cite{41}. Then, the Module 1 in VQABQ model, is our basic question generation module. In other words, in our model ,the only difference compared to \cite{41} is our Module 1, illustrated by Figure \ref{fig:figure2}.

\begin{table}
\centering
\begin{tabular}{|l|l|l|l|l|}
\hline
\multicolumn{5}{|c|}{Opend-Ended Case (Total: 142093 questions)}\\ 
\hline
& \begin{tabular}[c]{@{}l@{}}~0 BQ \\
(43\%)\end{tabular} & \begin{tabular}[c]{@{}l@{}}~~~1 BQ \\ (46.74\%)\end{tabular} & \begin{tabular}[c]{@{}l@{}}~~2 BQ \\ (5.44\%)\end{tabular} & \begin{tabular}[c]{@{}l@{}}~~3 BQ \\ (4.82\%)\end{tabular} \\
\hline
\# Q & 61100  & ~~66414  & ~~7730  & ~~6849 \\ 
\hline
\end{tabular}
\caption{We only show the open-ended case of VQA dataset \cite{4}, and "\# Q" denoted number of questions.}
\label{table:table2}
\end{table}

\begin{table}
\centering
\begin{tabular}{|l|l|l|l|}
\hline
\multicolumn{1}{|c|}{} & score1 & score2/score1 & score3/score2 \\ \hline
avg & ~~0.43  & ~~~~~~~0.49 & ~~~~~~~~0.73 \\ 
\hline
std & ~~0.31  & ~~~~~~~0.33 & ~~~~~~~~0.20 \\
\hline
\end{tabular}
\caption{"avg" denotes average and "std" denotes standard deviation.}
\label{table:table3}
\end{table}

\begin{table}
    \small
    \centering
    \begin{tabular}{ c | c c c c | c} 
      Task Type &    & \multicolumn{3}{c}{Open-Ended} &  \\ [0.5ex]
     \hline
      Test Set&  \multicolumn{4}{c}{dev} & std \\ [0.5ex]
     \hline
     Method & Num & Y/N & Other & All & All \\ [0.5ex] 
     \hline
     LSTM Q+I \cite{4} & 36.8 & 80.5 & 43.0 & 57.8 & 58.2  \\ 
     
     BOWIMG \cite{4} & 33.7 & 75.8 & 37.4 & 52.6 & - \\
     
     iBOWIMG \cite{53} & 35.0 & 76.6 & 42.6 & 55.7 & 55.9  \\
     
     DPPnet \cite{6} & 37.2 & 80.7 & 41.7 & 57.2 & 57.4  \\
     
     FDA \cite{55} & 36.2 & 81.1 & 45.8 & 59.2 & 59.5 \\
     
     SAN \cite{35} & 36.6 & 79.3 & 46.1 & 58.7 & 58.9  \\
     
     SMem \cite{56} & 37.3 & 80.9 & 43.1 & 58.0 & 58.2  \\
     
     DMN+ \cite{54} & 36.8 & 80.5 & 48.3 & 60.3 & 60.4 \\
     
     Refined-Neurons  \cite{57} & 36.4 & 78.4 & 46.3 & 58.4 & 58.4 \\
     
     QRU \cite{42} & 37.0 & 82.3 & 47.7 & 60.7 & 60.8 \\
     
     CoAtt+VGG \cite{41} & \textbf{38.4} & \textbf{79.6} & \textbf{49.1} & \textbf{60.5} & - \\
     
     CoAtt+ResNet \cite{41} & 38.7 & 79.7 & 51.7 & 61.8 & 62.1 \\
     \hline
     
   %  VQABQ+VGG(1)  & \textbf{38.5} & 78.8 & 46.4 & 58.9 & 59.2\\
     
%     Ours+VGG(2) [] & 38.2 & 78.2 & 45.9 & 58.3 & -\\
     
     Ours+VGG(1)  & 38.2 & 79.7 & 47.0 & 59.5 & -\\
    Ours+VGG(2)  & \textbf{38.4} & \textbf{79.7} & \textbf{49.1} & \textbf{60.5} & \textbf{60.3} \\

     \hline
    \end{tabular}
    \caption{Evaluation results on VQA dataset \cite{4}. "-" indicates the results are not available, and the Ours+VGG(1) and Ours+VGG(2) are the results by using different thresholds. Note that our VGGNet is same as CoAtt+VGG.}
\label{table:table4}
\end{table}

\begin{table}
    \small
    \centering
    \begin{tabular}{c | c c c c | c} 
      Task Type &    & \multicolumn{3}{c}{Open-Ended} & \\ [0.5ex]
     \hline
      Test Set&  \multicolumn{4}{c}{dev} & std \\ [0.5ex]
     \hline
     Method & Num & Y/N & Other & All & All \\ [0.5ex] 
     \hline
     
     LSTM Q+I \cite{55} & 36.46 & 80.87 & 43.40 & 58.02 & 58.18  \\
     
     CoAtt+VGG \cite{41} & \textbf{38.35} & \textbf{79.63} & 49.14 & \textbf{60.48} & \textbf{60.32} \\
     
%     CoAtt-a+ResNet [] & - & - & - & - & -  \\
     \hline
     
     %VQABQ+VGG(1)  & \textbf{38.5} & 78.8 & 46.4 & 58.9 & 59.2 \\
     
     Ours+VGG(2)  & \textbf{38.43} & \textbf{79.65} & 49.12 & \textbf{60.49} & \textbf{60.34} \\
     
%     Orus+ResNet [] & - & - & - & - & - \\

     \hline
    \end{tabular}
    \caption{Re-run evaluation results on VQA dataset \cite{4}. "-" indicates the results are not available. Note that the result of \cite{41} in Table \ref{table:table5} is lower than in Table \ref{table:table4}, and CoAtt+VGG is same as our VGGNet. According to the re-run results, our method has the higher accuracy, especially in the counting-type question. }
\label{table:table5}
\end{table}

\subsection{Evaluation Metrics}

\noindent

VQA dataset provides multiple-choice and open-ended task for evaluation. Regarding open-ended task, the answer can be any phrase or word. However, in multiple-choice task, an answer should be chosen from 18 candidate answers. For both cases, answers are evaluated by accuracy which can reflect human consensus. The accuracy is given by the following:
\begin{equation}
    Accuracy_{_{VQA}}=\frac{1}{N}\sum_{i=1}^{N}\min\left \{ \frac{\sum_{t\in T_{i}}\mathbb{I}[a_{i}=t]}{3},1 \right \}
\end{equation}
, where $N$ is the total number of examples, $\mathbb{I}[\cdot]$ denotes an indicator function, $a_{i}$ is the predicted answer and $T_{i}$ is an answer set of the $i^{th}$ example. That is, a predicted answer is considered as a correct one if at least 3 annotators agree with it, and the score depends on the total number of agreements when the predicted answer is not correct.

%COCO-QA dataset employs the classification accuracy and its relaxed version based on %the word similarity measure, Wu-Palmer Similarity (WUPS), and WUPS is given by the %following:
%\begin{equation}
%    \label{eq:magical}
%    \frac{1}{N}\sum_{i=1}^{N}\min\left \{ \prod_{a\in A_{i}}\max_{t\in T_{i}}\mu %(a,t),\prod_{t\in T_{i}}\max_{a\in A_{i}}\mu (a,t) \right \}
%\end{equation}
%, where $N$ is the total number of examples, $A_{i}$ is the predicted answer set and %$T_{i}$ is the ground truth answer set of the $i^{th}$ example. Note that $\mu %(\cdot,\cdot)$ denotes the thresholded WUPS between ground truth and prediction and %here we use two threshold values, 0.9 and 0.0, for evaluation. 

\begin{table*}
\centering
\begin{tabular}{|c|c|c|}
\hline
Main Question  & \multicolumn{1}{l|}{Corresponding Weight} & Basic Question \\ \hline
What type of computer is this?  & \begin{tabular}[c]{@{}c@{}}0.975978\\
0.000000\\
0.000000\end{tabular} & \begin{tabular}[c]{@{}c@{}}What type of computer is this?\\    
What is the players job in the foreground? \\   
How many light colored animals?\end{tabular} \\
\hline
Is this a farm? & \begin{tabular}[c]{@{}c@{}}0.975995\\
0.000000\\ 
0.000000\end{tabular} & \begin{tabular}[c]{@{}c@{}}Is this a farm?\\ 
What number is on the marker?\\ 
What picture is on the bag?\end{tabular}\\
\hline
What are these animals? & \begin{tabular}[c]{@{}c@{}}0.975976\\ 
0.000000\\ 
0.000000\end{tabular} & \begin{tabular}[c]{@{}c@{}}What are these animals? \\
Is there any food in this photo?\\
what website is there?\end{tabular}\\ 
\hline
Is the bed made?  & \begin{tabular}[c]{@{}c@{}}0.975996\\ 
0.000000\\ 
0.000000\end{tabular} & \begin{tabular}[c]{@{}c@{}}Is the bed made?\\ 
What storage is open?\\ 
Are both animals adult?\end{tabular}\\ 
\hline
Where is the baby?  & \begin{tabular}[c]{@{}c@{}}0.975993\\ 
0.000000\\ 
0.000000\end{tabular} & \begin{tabular}[c]{@{}c@{}}Where is the baby?\\ 
Are the cats facing each other?\\ 
How was their surfing run?\end{tabular}\\ 
\hline
What dessert is pictured on the plate? & \begin{tabular}[c]{@{}c@{}}0.975985\\ 0.000000\\ 
0.000000\end{tabular} & \begin{tabular}[c]{@{}c@{}}What dessert is pictured on the plate?\\ 
Is there anybody in the water?\\ 
Is this a home, apartment, or hotel ?\end{tabular}\\ 
\hline
\end{tabular}
\caption{Some failed examples about finding no basic question.}
\label{table:table6}
\end{table*}

% \begin{figure*}
% \begin{center}
% \fbox{\rule{0pt}{2in} \rule{.9\linewidth}{0pt}}
% \end{center}
%   \caption{Failed examples about finding no basic question.}
% \label{fig:short}
% \end{figure*}

\begin{figure*}
\begin{center}
 \includegraphics[scale = 0.4]{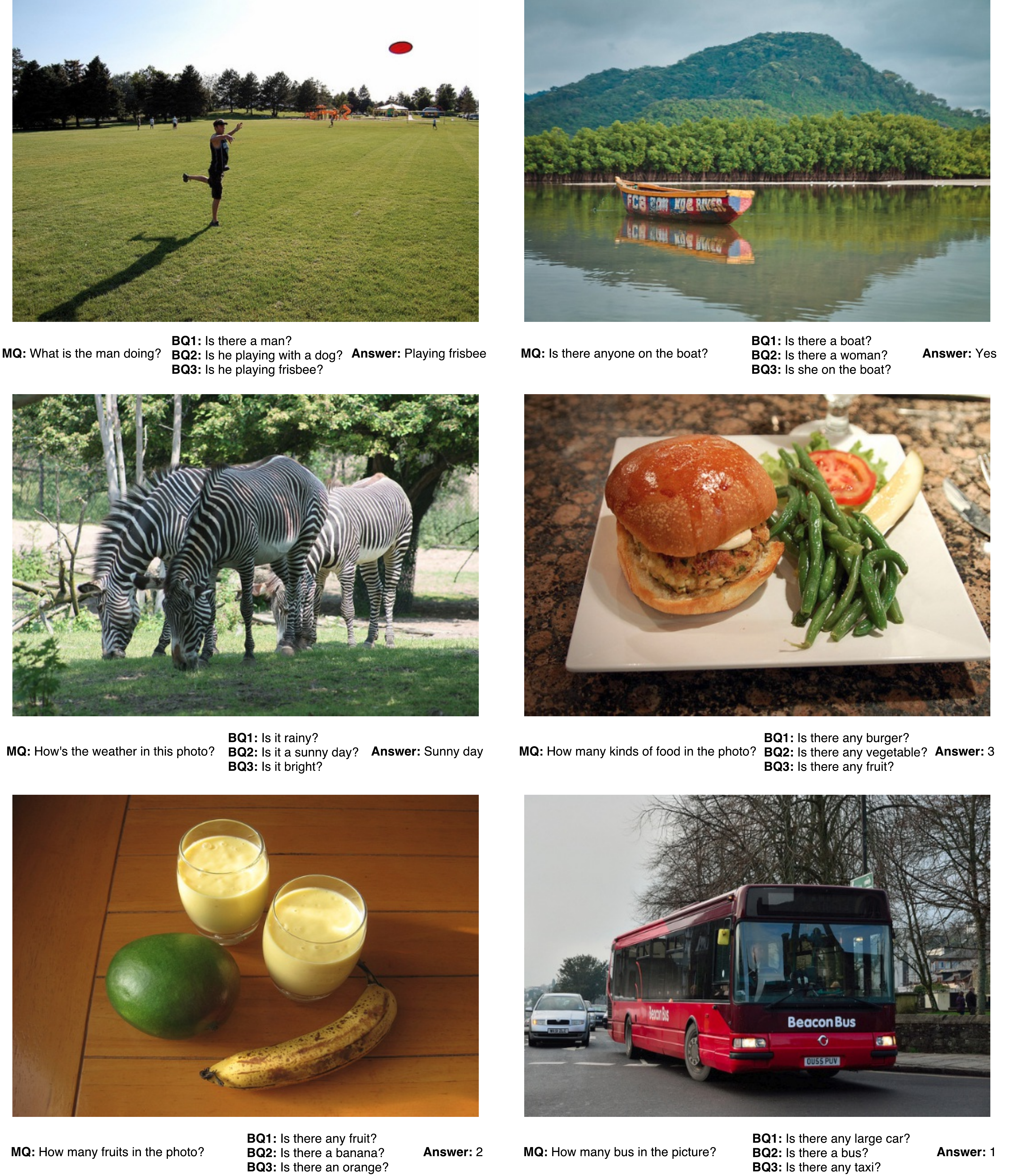}
%\fbox{\rule{0pt}{2in} \rule{.9\linewidth}{0pt}}
\end{center}
   \caption{Some future work examples.}
\label{fig:figure3}
\end{figure*}

\subsection{Results and Analysis}

Here, we describe our final results and analysis by the following parts:

\bigskip
\noindent
\textbf{Does Basic Question Help Accuracy ?}

The answer is yes. Here we only discuss the open-ended case. In our experiment, we use the $avg\pm std$, referring to Table \ref{table:table3}, to be the initial guess of proper thresholds of s1, s2 and s3, in Table \ref{table:table1}. We discover that when s1 = 0.43, s2 = 0.82 and s3 = 0.53, we can get the better utilization of BQ. The threshold, s1 = 0.43, can be consider as 43\% of testing questions from VQA dataset which cannot find the basic question, from the training and validation sets of VQA dataset, and only 57\% of testing questions can find the basic questions. Note that we combine the training and validation sets of VQA dataset to be our basic question dataset. Regarding s2 = 0.82, that means 82\% of those 57\% testing questions, i.e. 46.74\%, only can find 1 basic question, and 18\% of those 57\% testing questions, i.e. 10.26\%, can find at least 2 basic questions. Furthermore, s3 = 0.53 means that 53\% of those 10.26\% testing question, i.e. around 5.44\%, only can find 2 basic questions, and 47\% of those 10.26\% testing question, i.e. around 4.82\%, can find 3 basic questions. The above detail can be referred to Table \ref{table:table2}. 

Accordingly to the Table \ref{table:table2}, 43\% of testing questions from VQA dataset cannot find the proper basic questions from VQA training and validation datasets, and there are some failed examples about this case in Table \ref{table:table6}. We also discover that a lot of questions in VQA training and validation datasets are almost the same. This issue reduces the diversity of basic question dataset. Although we only have 57\% of testing questions can benefit from the basic questions, our method still can improve the state-of-the-art accuracy \cite{41} from 60.32\% to 60.34\%, referring to Table \ref{table:table4} and \ref{table:table5}. Then, we have 142093 testing questions, so that means the number of correctly answering questions of our method is more than state-of-the-art method 28 questions. In other words, if we have well enough basic question dataset, we can increase accuracy more, especially in the counting-type question, referring to Table \ref{table:table4} and \ref{table:table5}. Because the Co-Attention Mechanism is good at localizing, the counting-type question is improved more than others. So, based on our experiment, we can conclude that basic question can help accuracy obviously.

\bigskip

\noindent
\textbf{Comparison with State-of-the-art.}

Recently, \cite{41} proposed the Co-Attention Mechanism in VQA and got the state-of-the-art accuracy. However, when we use their code and the same setup mentioned in their paper to re-run the experiment, we cannot get the same accuracy reported in their work. The re-run results are presented in Table \ref{table:table5}. So, under the fair conditions, our method is competitive compared to the state-of-the-art.

%\bigskip

%\noindent
%\textbf{Ablation Study.}

%conclusion of this work and some future work based on this paper.Here we %will describe the conclusion of this work and some future work based this %paper.

%\bigskip
%\noindent
%\textbf{Qualitative Results.}

%Here we will describe the conclusion of this work and some future work %based on this paper.Here we will describe the conclusion of this 

%------------------------------------------------------------------------
\section{Conclusion and Future Work}

In this paper, we propose a VQABQ model for visual question answering. The VQABQ model has two main modules, Basic Question Generation Module and Co-Attention VQA Module. The former one can generate the basic questions for the query question, and the latter one can take the image , basic and query question as input and then output the text-based answer of the query question. According to the Section 6.4, because the basic question dataset generated from VQA dataset is not well enough, we only have the 57\% of all testing questions can benefit from the basic questions. However, we still can increase 28 correctly answering questions compared to the state-of-the-art. We believe that if our basic question dataset is well enough, the increment of accuracy will be much more.

According to the previous state-of-the-art methods in VQA, they all got the highest accuracy in the Yes/No-type question. So, how to effectively only exploit the Yes/No-type basic questions to do VQA will be an interesting work, illustrated by Figure \ref{fig:figure3}. Also, how to generate other specific type of basic questions based on the query question and how to do better combination of visual and textual features in order to decrease the semantic inconsistency? The above future works will be our next research focus.

%-------------------------------------------------------------------------

\section*{Acknowledgements} 

This work is supported by competitive research funding from King Abdullah University of Science and Technology (KAUST). Also, we would like to acknowledge Fabian Caba, Humam Alwassel and Adel Bibi. They always can provide us helpful discussion about this work.

{\small
\bibliographystyle{ieee}
\bibliography{iccv2017ref}

\begin{thebibliography}{10}\itemsep=-1pt

\bibitem{4}
S.~Antol, A.~Agrawal, J.~Lu, M.~Mitchell, D.~Batra, C.~Lawrence~Zitnick, and
  D.~Parikh.
\newblock Vqa: Visual question answering.
\newblock In {\em Proceedings of the IEEE International Conference on Computer
  Vision}, pages 2425--2433, 2015.

\bibitem{15}
K.~Chen, J.~Wang, L.-C. Chen, H.~Gao, W.~Xu, and R.~Nevatia.
\newblock Abc-cnn: An attention based convolutional neural network for visual
  question answering.
\newblock {\em arXiv preprint arXiv:1511.05960}, 2015.

\bibitem{46}
K.~Cho, B.~Van~Merri{\"e}nboer, C.~Gulcehre, D.~Bahdanau, F.~Bougares,
  H.~Schwenk, and Y.~Bengio.
\newblock Learning phrase representations using rnn encoder-decoder for
  statistical machine translation.
\newblock {\em arXiv preprint arXiv:1406.1078}, 2014.

\bibitem{10}
J.~Chung, C.~Gulcehre, K.~Cho, and Y.~Bengio.
\newblock Empirical evaluation of gated recurrent neural networks on sequence
  modeling.
\newblock {\em arXiv preprint arXiv:1412.3555}, 2014.

\bibitem{48}
H.~Fang, S.~Gupta, F.~Iandola, R.~K. Srivastava, L.~Deng, P.~Doll{\'a}r,
  J.~Gao, X.~He, M.~Mitchell, J.~C. Platt, et~al.
\newblock From captions to visual concepts and back.
\newblock In {\em Proceedings of the IEEE Conference on Computer Vision and
  Pattern Recognition}, pages 1473--1482, 2015.

\bibitem{33}
H.~Gao, J.~Mao, J.~Zhou, Z.~Huang, L.~Wang, and W.~Xu.
\newblock Are you talking to a machine? dataset and methods for multilingual
  image question.
\newblock In {\em Advances in Neural Information Processing Systems}, pages
  2296--2304, 2015.

\bibitem{44}
S.~Hochreiter and J.~Schmidhuber.
\newblock Long short-term memory.
\newblock {\em Neural computation}, 9(8):1735--1780, 1997.

\bibitem{55}
I.~Ilievski, S.~Yan, and J.~Feng.
\newblock A focused dynamic attention model for visual question answering.
\newblock {\em arXiv preprint arXiv:1604.01485}, 2016.

\bibitem{18}
K.~Kafle and C.~Kanan.
\newblock Answer-type prediction for visual question answering.
\newblock In {\em Proceedings of the IEEE Conference on Computer Vision and
  Pattern Recognition}, pages 4976--4984, 2016.

\bibitem{27}
A.~Karpathy and L.~Fei-Fei.
\newblock Deep visual-semantic alignments for generating image descriptions.
\newblock In {\em Proceedings of the IEEE Conference on Computer Vision and
  Pattern Recognition}, pages 3128--3137, 2015.

\bibitem{43}
R.~Kiros, Y.~Zhu, R.~R. Salakhutdinov, R.~Zemel, R.~Urtasun, A.~Torralba, and
  S.~Fidler.
\newblock Skip-thought vectors.
\newblock In {\em NIPS}, pages 3294--3302, 2015.

\bibitem{42}
R.~Li and J.~Jia.
\newblock Visual question answering with question representation update (qru).
\newblock In {\em NIPS}, pages 4655--4663, 2016.

\bibitem{51}
T.-Y. Lin, M.~Maire, S.~Belongie, J.~Hays, P.~Perona, D.~Ramanan,
  P.~Doll{\'a}r, and C.~L. Zitnick.
\newblock Microsoft coco: Common objects in context.
\newblock In {\em ECCV}, pages 740--755. Springer, 2014.

\bibitem{41}
J.~Lu, J.~Yang, D.~Batra, and D.~Parikh.
\newblock Hierarchical question-image co-attention for visual question
  answering.
\newblock In {\em NIPS}, pages 289--297, 2016.

\bibitem{30}
L.~Ma, Z.~Lu, and H.~Li.
\newblock Learning to answer questions from image using convolutional neural
  network.
\newblock {\em arXiv preprint arXiv:1506.00333}, 2015.

\bibitem{14}
L.~Ma, Z.~Lu, and H.~Li.
\newblock Learning to answer questions from image using convolutional neural
  network.
\newblock In {\em AAAI}, page~16, 2016.

\bibitem{2}
M.~Malinowski and M.~Fritz.
\newblock A multi-world approach to question answering about real-world scenes
  based on uncertain input.
\newblock In {\em Advances in Neural Information Processing Systems}, pages
  1682--1690, 2014.

\bibitem{9}
M.~Malinowski, M.~Rohrbach, and M.~Fritz.
\newblock Ask your neurons: A neural-based approach to answering questions
  about images.
\newblock In {\em Proceedings of the IEEE International Conference on Computer
  Vision}, pages 1--9, 2015.

\bibitem{57}
M.~Malinowski, M.~Rohrbach, and M.~Fritz.
\newblock Ask your neurons: A deep learning approach to visual question
  answering.
\newblock {\em International Journal of Computer Vision (IJCV)}, 2017.
\newblock to appear.

\bibitem{47}
T.~Mikolov, I.~Sutskever, K.~Chen, G.~S. Corrado, and J.~Dean.
\newblock Distributed representations of words and phrases and their
  compositionality.
\newblock In {\em NIPS}, pages 3111--3119, 2013.

\bibitem{6}
H.~Noh, P.~Hongsuck~Seo, and B.~Han.
\newblock Image question answering using convolutional neural network with
  dynamic parameter prediction.
\newblock In {\em Proceedings of the IEEE Conference on Computer Vision and
  Pattern Recognition}, pages 30--38, 2016.

\bibitem{49}
K.~Papineni, S.~Roukos, T.~Ward, and W.-J. Zhu.
\newblock Bleu: a method for automatic evaluation of machine translation.
\newblock In {\em Proceedings of the 40th annual meeting on association for
  computational linguistics}, pages 311--318. Association for Computational
  Linguistics, 2002.

\bibitem{11}
J.~Pennington, R.~Socher, and C.~D. Manning.
\newblock Glove: Global vectors for word representation.
\newblock In {\em EMNLP}, volume~14, pages 1532--1543, 2014.

\bibitem{12}
M.~Ren, R.~Kiros, and R.~Zemel.
\newblock Exploring models and data for image question answering.
\newblock In {\em Advances in Neural Information Processing Systems}, pages
  2953--2961, 2015.

\bibitem{32}
M.~Ren, R.~Kiros, and R.~Zemel.
\newblock Exploring models and data for image question answering.
\newblock In {\em Advances in Neural Information Processing Systems}, pages
  2953--2961, 2015.

\bibitem{13}
K.~J. Shih, S.~Singh, and D.~Hoiem.
\newblock Where to look: Focus regions for visual question answering.
\newblock In {\em Proceedings of the IEEE Conference on Computer Vision and
  Pattern Recognition}, pages 4613--4621, 2016.

\bibitem{45}
I.~Sutskever, O.~Vinyals, and Q.~V. Le.
\newblock Sequence to sequence learning with neural networks.
\newblock In {\em NIPS}, pages 3104--3112, 2014.

\bibitem{28}
O.~Vinyals, A.~Toshev, S.~Bengio, and D.~Erhan.
\newblock Show and tell: A neural image caption generator.
\newblock In {\em Proceedings of the IEEE Conference on Computer Vision and
  Pattern Recognition}, pages 3156--3164, 2015.

\bibitem{37}
Q.~Wu, P.~Wang, C.~Shen, A.~Dick, and A.~van~den Hengel.
\newblock Ask me anything: Free-form visual question answering based on
  knowledge from external sources.
\newblock In {\em CVPR}, pages 4622--4630, 2016.

\bibitem{54}
C.~Xiong, S.~Merity, and R.~Socher.
\newblock Dynamic memory networks for visual and textual question answering.
\newblock {\em arXiv}, 1603, 2016.

\bibitem{56}
H.~Xu and K.~Saenko.
\newblock Ask, attend and answer: Exploring question-guided spatial attention
  for visual question answering.
\newblock In {\em ECCV}, pages 451--466. Springer, 2016.

\bibitem{1}
K.~Xu, J.~Ba, R.~Kiros, K.~Cho, A.~C. Courville, R.~Salakhutdinov, R.~S. Zemel,
  and Y.~Bengio.
\newblock Show, attend and tell: Neural image caption generation with visual
  attention.
\newblock In {\em ICML}, volume~14, pages 77--81, 2015.

\bibitem{35}
Z.~Yang, X.~He, J.~Gao, L.~Deng, and A.~Smola.
\newblock Stacked attention networks for image question answering.
\newblock In {\em Proceedings of the IEEE Conference on Computer Vision and
  Pattern Recognition}, pages 21--29, 2016.

\bibitem{50}
J.~Yao, S.~Fidler, and R.~Urtasun.
\newblock Describing the scene as a whole: Joint object detection, scene
  classification and semantic segmentation.
\newblock In {\em CVPR 2012}, pages 702--709. IEEE, 2012.

\bibitem{53}
B.~Zhou, Y.~Tian, S.~Sukhbaatar, A.~Szlam, and R.~Fergus.
\newblock Simple baseline for visual question answering.
\newblock {\em arXiv preprint arXiv:1512.02167}, 2015.

\bibitem{34}
Y.~Zhu, O.~Groth, M.~Bernstein, and L.~Fei-Fei.
\newblock Visual7w: Grounded question answering in images.
\newblock In {\em Proceedings of the IEEE Conference on Computer Vision and
  Pattern Recognition}, pages 4995--5004, 2016.

\end{thebibliography}
}

\end{document}